\documentclass[11pt,a4paper]{article}
\usepackage[hyperref]{acl2018}
\usepackage{times}
\usepackage{latexsym}
\usepackage{url}

\usepackage{cleveref}
\crefformat{section}{\S#2#1#3} 
\crefformat{subsection}{\S#2#1#3}
\crefformat{subsubsection}{\S#2#1#3}

\usepackage{algorithm}
\usepackage{algorithmic}
\usepackage{enumerate}
\usepackage{amsmath} 
\usepackage{amssymb}  

\usepackage{booktabs}
\usepackage{multirow}
\usepackage{float}
\usepackage{graphicx}
\usepackage{footnote}
\usepackage{tablefootnote}
\usepackage{subcaption}

\newcommand*{\affaddr}[1]{#1} 
\newcommand*{\affmark}[1][*]{\textsuperscript{#1}}
\newcommand*{\email}[1]{\texttt{#1}}

\aclfinalcopy 
\def\aclpaperid{555} 

\title{Exploiting Document Knowledge \\ for Aspect-level Sentiment Classification}

\author{Ruidan He\affmark[\dag\ddag], Wee Sun Lee\affmark[\dag], Hwee Tou Ng\affmark[\dag], \and Daniel Dahlmeier\affmark[\ddag]\\
\affaddr{\affmark[\dag]Department of Computer Science, National University of Singapore}\\
\affaddr{\affmark[\ddag]SAP Innovation Center Singapore}\\
\email{\affmark[\dag]\{ruidanhe,leews,nght\}@comp.nus.edu.sg}\\
\email{\affmark[\ddag]d.dahlmeier@sap.com}%
}

\date{}

\begin{document}
\maketitle
\begin{abstract}

Attention-based long short-term memory (LSTM) networks have proven to be useful in aspect-level sentiment classification. However, due to the difficulties in annotating aspect-level data, existing public datasets for this task are all relatively small, which largely limits the effectiveness of those neural models. 
In this paper, we explore two approaches that transfer knowledge from document-level data, which is much less expensive to obtain, to improve the performance of aspect-level sentiment classification. We demonstrate the effectiveness of our approaches on 4 public datasets from SemEval 2014, 2015, and 2016, and we show that attention-based LSTM benefits from document-level knowledge in multiple ways.
\end{abstract}

\section{Introduction}

Given a sentence and an opinion target (also called an aspect term) occurring in the sentence, aspect-level sentiment classification aims to determine the sentiment polarity in the sentence towards the opinion target. An opinion target or target for short refers to a word or a phrase describing an aspect of an entity. For example, in the sentence ``\emph{This little place has a cute interior decor but the prices are quite expensive}'', the targets are \emph{interior decor} and \emph{prices}, and they are associated with positive and negative sentiment respectively.

A sentence may contain multiple sentiment-target pairs, thus one challenge is to separate different opinion contexts for different targets. For this purpose, state-of-the-art neural methods~\cite{Wang:16,Liu:17,Chen:17} adopt attention-based LSTM networks, where the LSTM aims to capture sequential patterns and the attention mechanism aims to emphasize target-specific contexts for encoding sentence representations. Typically, LSTMs only show their potential when trained on large datasets. However, aspect-level training data requires the annotation of all opinion targets in a sentence, which is costly to obtain in practice. As such, existing public aspect-level datasets are all relatively small. Insufficient training data limits the effectiveness of neural models.  

Despite the lack of aspect-level labeled data, enormous document-level labeled data are easily accessible online such as Amazon reviews. These reviews contain substantial linguistic patterns and come with sentiment labels naturally. In this paper,  we hypothesize that aspect-level sentiment classification can be improved by employing knowledge gained from document-level sentiment classification. Specifically, we explore two transfer methods to incorporate this sort of knowledge -- pretraining and multi-task learning. In our experiments, we find that both methods are helpful and combining them achieves significant improvements over attention-based LSTM models trained only on aspect-level data. We also illustrate by examples that additional knowledge from document-level data is beneficial in multiple ways. 
Our source code can be obtained from \url{https://github.com/ruidan/Aspect-level-sentiment}.

\section{Related Work}
Various neural
models~\cite{Dong:14,Nguyen:15,Vo:15,Tang:16a,Tang:16b,Wang:16,zhang:16,Liu:17,Chen:17} have been proposed for aspect-level sentiment classification. The main idea behind these works is to develop neural architectures that are able to learn continuous features and capture the intricate relation between a target and context words. However, to sufficiently train these models, substantial aspect-level annotated data is required, which is expensive to obtain in practice.

We explore both pretraining and multi-task learning for transferring knowledge from document level to aspect level. Both methods are widely studied in the literature. Pretraining is a common technique used in computer vision where low-level neural layers can be usefully transferred to different tasks~\cite{Krizhevsky:12,Zeiler:14}. In natural language processing (NLP), some efforts have been initiated on pretraining LSTMs~\cite{Dai:15,Zoph:16,Prajit:17} for sequence-to-sequence models in both supervised and unsupervised settings, where promising results have been obtained. On the other hand, multi-task learning simultaneously trains on samples in multiple tasks with a combined objective~\cite{Collobert:08,Luong:15a,Liu:16}, which has improved model generalization ability in certain cases. In the work of Mou et al.~\shortcite{Mou:16}, the authors investigated the transferability of neural models in NLP applications with extensive experiments, showing that transferability largely depends on the semantic relatedness of the source and target tasks. For our problem, we hypothesize that aspect-level sentiment classification can be improved by employing knowledge gained from document-level sentiment classification, as these two tasks are highly related semantically.

\section{Models}
\subsection{Attention-based LSTM}
We first describe a conventional implementation of an attention-based LSTM model for this task. We use it as a baseline model and extend it with pretraining and multi-task learning approaches for incorporating document-level knowledge. 

The inputs are a sentence $s = (w_1, w_2, ..., w_n)$ consisting of $n$ words, and an opinion target $x = (x_1, x_2, ..., x_m)$ occurring in the sentence consisting of a subsequence of $m$ words from $s$. Each word is associated with a continuous word embedding $\mathbf{e}_w$ ~\cite{Mikolov:13} from an embedding matrix $\mathbf{E} \in \mathbb{R}^{V \times d}$, where $V$ is the vocabulary size and $d$ is the embedding dimension.

LSTM is used to capture sequential information, and outputs a sequence of hidden vectors:
\begin{equation}
[\mathbf{h}_1, ..., \mathbf{h}_n] = \text{LSTM}([\mathbf{e}_{w_1}, ..., \mathbf{e}_{w_n}], \theta_{lstm})
\end{equation}
An attention layer assigns a weight $\alpha_i$ to each word in the sentence. The final target-specific representation of the sentence $s$ is then given by:
\begin{equation}
\mathbf{z} = \sum_{i=1}^{n} \alpha_i \mathbf{h}_i
\end{equation}
And $\alpha_i$ is computed as follows:
\begin{align}
&\alpha_i = \frac{\exp(\beta_i)}{\sum_{j=1}^{n}\exp(\beta_j)} \\
&\beta_i = f_{score}(\mathbf{h}_i, \mathbf{t}) = tanh(\mathbf{h}_i^T \mathbf{W}_a \mathbf{t})\\
&\mathbf{t} = \frac{1}{m} \sum_{i=1}^{m}\mathbf{e}_{x_i} \label{target_rep}
\end{align}
where $\mathbf{t}$ is the target representation computed as the averaged word embedding of the target. $f_{score}$ is a content-based function that captures the semantic association between a word and the target, for which we adopt the formulation used in~\cite{Luong:15b,He:17} with parameter matrix $\mathbf{W}_a \in \mathbb{R}^{d \times d}$. 

The sentence representation $\mathbf{z}$ is fed into an output layer to predict the probability distribution $\mathbf{p}$ over sentiment labels on the target:
\begin{equation}
\mathbf{p}=softmax(\mathbf{W}_o \mathbf{z} + \mathbf{b}_o) \label{output layer}
\end{equation}
We refer to this baseline model as LSTM+ATT. It is trained via cross entropy minimization:
\begin{equation}
J = -\sum_{i \in D} \log \mathbf{p}_i(c_i) \label{objective}
\end{equation}
where $D$ denotes the overall training corpus, $c_i$ denotes the true label for sample $i$, and $\mathbf{p}_i(c_i)$ denotes the probability of the true label.

\subsection{Transfer Approaches}
LSTM+ATT is used as our aspect-level model with parameter set $\theta_{aspect} = \{ \mathbf{E}, \theta_{lstm}, \mathbf{W}_a, \mathbf{W}_o, \mathbf{b}_o \}$. We also build a standard LSTM-based classifier based on document-level training examples. This network is the same as the LSTM+ATT apart from the lack of the attention layer. The training objective is also cross entropy minimization as shown in equation (\ref{objective}) and the parameter set is $\theta_{doc} = \{ \mathbf{E}^{\prime}, \theta_{lstm}^{\prime}, \mathbf{W}_o^{\prime}, \mathbf{b}_o^{\prime} \}$.
\medskip

\noindent\textbf{Pretraining} (PRET): In this setting, we first train on document-level examples. The last hidden vector from the LSTM outputs is used as the document representation.
We initialize the relevant parameters $\mathbf{E}, \theta_{lstm}, \mathbf{W}_o, \mathbf{b}_o$ of LSTM+ATT with the pretrained weights, and train it on aspect-level examples to fine tune those weights and learn $\mathbf{W}_a$ which is randomly initialized.
\medskip

\noindent\textbf{Multi-task Learning} (MULT): This approach simultaneously trains two tasks -- document-level and aspect-level classification. In this setting, the embedding layer ($\mathbf{E}$) and the LSTM layer ($\theta_{lstm}$) are shared by both tasks, and a document is represented as the mean vector over LSTM outputs.  The other parameters are task-specific. The overall loss function is then given by:
\begin{equation}
L = J + \lambda U \label{overall objective}
\end{equation}
where $U$ is the loss function of document-level classification. $\lambda \in (0,1)$ is a hyperparameter that controls the weight of $U$.
\medskip

\noindent\textbf{Combined} (PRET+MULT): In this setting, we first perform PRET on document-level examples. We use the pretrained weights for parameter initialization for both aspect-level model and document-level model, and then perform MULT as discussed above.

\renewcommand{\arraystretch}{1.1}
\begin{table}[t]
\centering
\small
\scalebox{0.8}{
\begin{tabular}{llccc}
\toprule
& {Dataset} & Pos & Neg & Neu \\
\hline
\multirow{2}{*}{D1} & Restaurant14-Train & 2164 & 807 & 637 \\
& Restaurant14-Test & 728 & 196 & 196 \\
\hline
\multirow{2}{*}{D2} & Laptop14-Train & 994 & 870 & 464 \\
& Laptop14-Test & 341 & 128 & 169 \\
\hline
\multirow{2}{*}{D3} & Restaurant15-Train & 1178 & 382 & 50 \\
& Restaurant15-Test & 439 & 328 & 35 \\
\hline
\multirow{2}{*}{D4} & Restaurant16-Train & 1620 & 709 & 88 \\
& Restaurant16-Test & 597 & 190 & 38 \\
\bottomrule
\end{tabular}}
\caption{Dataset description.}\label{data statistics}
\end{table}

\begin{table*}[t]
\centering
\small
\scalebox{0.8}{
\begin{tabular}{lcccccccc}
\toprule 
\multirow{ 2}{*}{Methods} & \multicolumn{2}{c}{D1} & \multicolumn{2}{c}{D2} & \multicolumn{2}{c}{D3} & \multicolumn{2}{c}{D4}\\\cline{2-9}
&Acc. &Macro-F1 &Acc. &Macro-F1 &Acc. &Macro-F1 &Acc. &Macro-F1\\\hline
Tang et al. (\citeyear{Tang:16a}) &75.37 &64.51 &68.25 &65.96 &76.39 &58.70 &82.16 &54.21 \\
Wang et al. (\citeyear{Wang:16}) &78.60 &67.02 &68.88 &63.93 &78.48 &62.84 &83.77 &61.71 \\
Tang et al. (\citeyear{Tang:16b}) &76.87 &66.40 &68.91 &62.79 &77.89 &59.52 &83.04 &57.91 \\
Chen et al. (\citeyear{Chen:17}) &78.48 &68.54  &\bf{72.08} &68.43 &79.98 &60.57 &83.88 &62.14
\\\hline
LSTM &75.23 &64.21 &66.79 &64.02 &75.28 &54.10 &81.95 &58.11 \\
LSTM+ATT &76.83 &66.48 &68.07 &64.82 &77.38 &60.52 &82.73 &59.12 \\
Ours: PRET &78.28 &68.55 &71.32 &\bf{68.53} &80.67 &68.31 &84.87 &\bf{70.73} \\
Ours: MULT &77.41 &66.68 &68.65 &64.57 &81.05 &65.69 &83.27 &64.56 \\
Ours: PRET+MULT &\bf{79.11} &\bf{69.73$^*$} &71.15 &67.46 &\bf{81.30}$^*$ &\bf{68.74}$^*$ &\bf{85.58}$^*$ &69.76$^*$ \\
\bottomrule
\end{tabular}}
\caption{Average accuracies and Macro-F1 scores over 5 runs with random initialization. The best results are in bold. $^*$ indicates that PRET+MULT is significantly better than Tang et al. (\citeyear{Tang:16a}), Wang et al. (\citeyear{Wang:16}), Tang et al. (\citeyear{Tang:16b}), Chen et al. (\citeyear{Chen:17}), LSTM, and LSTM+ATT with $p < 0.05$ according to one-tailed unpaired t-test.}\label{model comparison}
\end{table*}

\begin{table*}[t]
\centering
\small
\scalebox{0.8}{
\begin{tabular}{lcccccccc}
\toprule 
\multirow{ 2}{*}{Settings} & \multicolumn{2}{c}{D1} & \multicolumn{2}{c}{D2} & \multicolumn{2}{c}{D3} & \multicolumn{2}{c}{D4}\\\cline{2-9}
&Acc. &Macro-F1 &Acc. &Macro-F1 &Acc. &Macro-F1 &Acc. &Macro-F1\\\hline
LSTM only &78.09 &67.85 &71.04 &66.80 &78.95 &65.30 &83.85 &67.11 \\
Embeddings only &77.12 &67.19 &69.12 &65.06 &80.13 &67.04 &84.12 &70.11 \\
Output layer only &76.88 &66.81 &69.63 &66.07 &78.30 &64.49 &82.55 &62.83 \\\hline
Without LSTM  &77.45 &67.25 &69.82 &66.63 &80.27 &68.02 &84.80 &70.27 \\
Without embeddings &77.97 &67.96 &70.59 &67.16 &79.08 &65.56 &83.94 &68.79 \\
Without output layer  &78.36 &68.06 &71.10 &67.87 &80.82 &67.68 &84.71 &70.48 \\
\bottomrule
\end{tabular}}
\caption{PRET with different transferred layers. Averaged results over 5 runs are reported.}\label{ablation test}
\end{table*}

\section{Experiments}
\subsection{Datasets and Experimental Settings}
We run experiments on four benchmark aspect-level datasets, taken from SemEval 2014~\cite{Pontiki:14}, SemEval 2015~\cite{Pontiki:15}, and SemEval 2016~\cite{Pontiki:16}. Following previous work~\cite{Tang:16b,Wang:16}, we remove samples with \emph{conflicting} polarities in all datasets\footnote{We remove samples in the 2015/6 datasets if an opinion target is associated with different sentiment polarities.}. Statistics of the resulting datasets are presented in Table~\ref{data statistics}.

We derived two document-level datasets from Yelp2014~\cite{tang:15} and the Amazon Electronics dataset~\cite{Mcauley:15} respectively. The original reviews were rated on a 5-point scale. We consider 3-class classification and thus label reviews with rating $<3$, $>3$, and $=3$ as negative, positive, and neutral respectively. Each sampled dataset contains 30k instances with exactly balanced class labels. We pair up an aspect-level dataset and a document-level dataset when they are from a similar domain -- the Yelp dataset is used by D1, D3, and D4 for PRET and MULT, and the Electronics dataset is only used by D2. 

In all experiments, 300-dimension GloVe vectors~\cite{Pennington:14} are used to initialize $\mathbf{E}$ and $\mathbf{E}^\prime$ when pretraining is not conducted for weight initialization. These vectors are also used for initializing $\mathbf{E}^\prime$ in the pretraining phase. Values for hyperparameters are obtained from experiments on development sets. We randomly sample 20\% of the original training data from the aspect-level dataset as the development set and only use the remaining 80\% for training. For all experiments, the dimension of LSTM hidden vectors is set to 300, $\lambda$ is set to 0.1, and we use dropout with probability 0.5 on sentence/document representations before the output layer. We use RMSProp as the optimizer with the decay rate set to 0.9 and the base learning rate set to 0.001. The mini-batch size is set to 32.

\subsection{Model Comparison}
Table \ref{model comparison} shows the results of LSTM, LSTM+ATT, PRET, MULT, PRET+MULT, and four representative prior works~\cite{Tang:16a,Tang:16b,Wang:16,Chen:17}.  Significance tests are conducted for testing the robustness of methods under random parameter initialization.
Both accuracy and macro-F1 are used for evaluation as label distribution is unbalanced. The reported numbers are obtained as the average value over 5 runs with random initialization for each method. 

We observe that PRET is very helpful, and consistently gives a 1--3\% increase in accuracy over LSTM+ATT across all datasets. The improvements in macro-F1 scores are even more, especially on D3 and D4 where the labels are extremely unbalanced. 
MULT gives similar performance as LSTM+ATT on D1 and D2, but improvements can be clearly observed for D3 and D4. The combination (PRET+MULT) overall yields better results.

There are two main reasons why the improvements of macro-F1 scores are more significant on D3 and D4 than on D1: (1) D1 has much more neutral examples in the training set. A classifier without any external knowledge might still be able to learn some neutral-related features on D1 but it is very hard to learn from D3 and D4. (2) The numbers of neutral examples in the test sets of D3 and D4 are very small. Thus, the precision and recall on neutral class will be largely affected by even a small prediction difference (e.g., with 5 more neutral examples correctly identified, recall is increased by more than 10\% on both datasets). As a result, the macro-F1 scores on D3 and D4 are affected more.

\subsection{Ablation Tests}
Table \ref{model comparison} indicates that a large percentage of the performance gain comes from PRET. To better understand the transfer effects of different layers -- embedding layer ($\mathbf{E}$), LSTM layer ($\theta_{lstm}$), and output layer ($\mathbf{W}_o, \mathbf{b}_o$) -- we conduct ablation tests on PRET with different layers transfered from the document-level model to the aspect-level model. Results are presented in Table \ref{ablation test}. ``LSTM only'' denotes the setting where only the LSTM layer is transferred, and ``Without LSTM'' denotes the setting where only the embedding and output layers are transferred (excluding the LSTM layer). The key observations are: (1) Transfer is helpful in all settings. Improvements over LSTM+ATT are observed even when only one layer is transferred. (2) Overall, transfers of the LSTM and embedding layer are more useful than the output layer. This is what we expect, since the output layer is normally more task-specific. 
(3) Transfer of the embedding layer is more helpful on D3 and D4. One possible explanation is that the label distribution is extremely unbalanced on these two datasets. Sentiment information is not adequately captured by GloVe word embeddings. Therefore, with a small number of training examples in the negative and neutral classes, the embeddings trained by aspect-level classification still do not effectively capture the true semantics of the relevant opinion words. Transfer of the embedding layer can greatly help in this case.

\subsection{Analysis}

\begin{figure}[t]
  \centering
  \begin{minipage}[b]{0.37\textwidth}
    \includegraphics[width=0.85\textwidth]{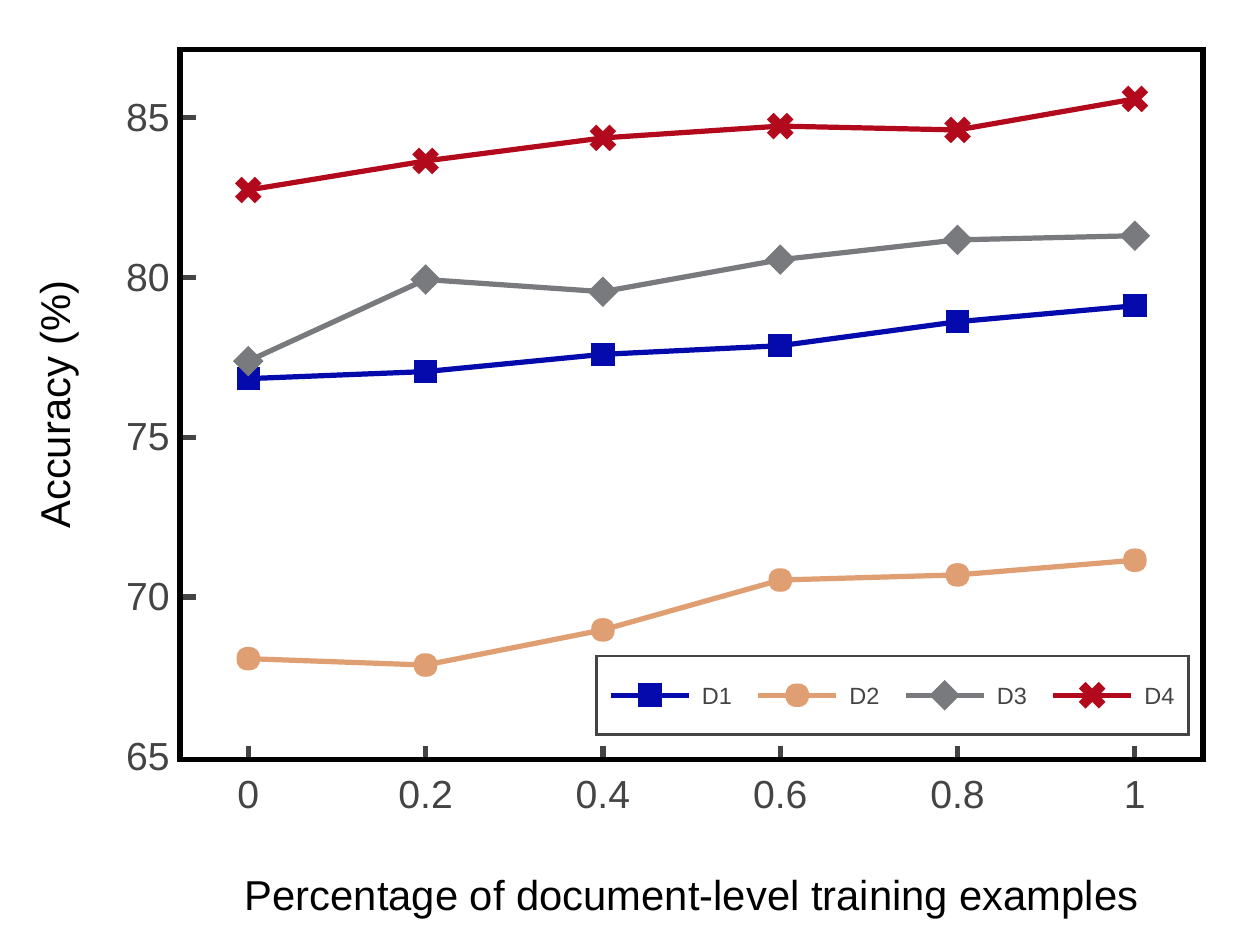}
  \end{minipage}
  \hfill
  \begin{minipage}[b]{0.37\textwidth}
    \includegraphics[width=0.85\textwidth]{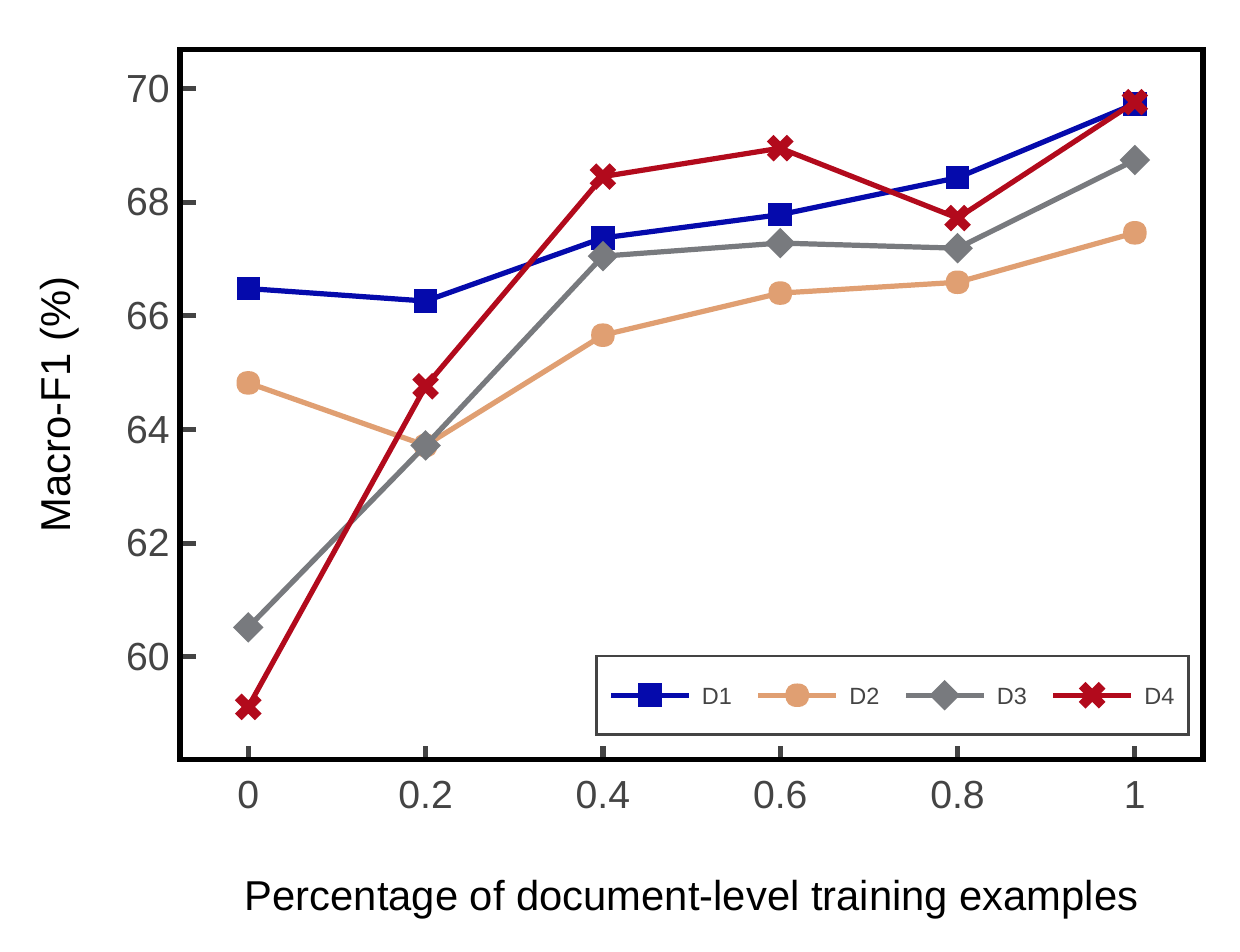}
  \end{minipage}
  \caption{Results of PRET+MULT vs. percentage of document-level training data.}
    \label{percentage}
\end{figure}

To show that aspect-level classification indeed benefits from document-level knowledge, we conduct experiments to vary the percentage of document-level training examples from 0.0 to 1.0 for PRET+MULT. The changes of accuracies and macro-F1 scores on the four datasets are shown in Figure~\ref{percentage}. The improvements on accuracies with increasing number of document examples are stable across all datasets. For macro-F1 scores, the improvements on D1 and D2 are stable. We observe sharp increases in the macro-F1 scores of D3 and D4 when changing the percentage from 0 to 0.4. This may be related to their extremely unbalanced label distribution. In such cases, with the knowledge gained from a small number of balanced document-level examples, aspect-level predictions on neutral examples can be significantly improved. 

To better understand in which conditions the proposed method is helpful, we analyze a subset of test examples that are correctly classified by PRET+MULT but are misclassified by LSTM+ATT. We find that the benefits brought by document-level knowledge are typically shown in four ways.

First of all, to our surprise, LSTM+ATT made obvious mistakes on some instances with common opinion words. Below are two examples where the target is enclosed in [] with its true sentiment indicated in the subscript:

1. \emph{``I was highly disappointed in the [food]$_{neg}$.''}

2. \emph{``This particular location certainly uses substandard [meats]$_{neg}$.''}

In the above examples, LSTM+ATT does attend to the right opinion words, but makes the wrong predictions. One possible reason is that the word embeddings from GloVe without PRET do not effectively capture sentiment information, while the aspect-level training samples are not sufficient to capture that for certain words. PRET+MULT eliminates this kind of errors.

Another finding is that our method helps to better capture domain-specific opinion words due to additional knowledge from documents that are from a similar domain:

1. \emph{``The smaller [size]$_{pos}$ was a bonus because of space restrictions.''}

2. \emph{``The [price]$_{pos}$ is 200 dollars down.''}

LSTM+ATT attends on \emph{smaller} correctly for the first example but makes the wrong prediction as \emph{smaller} can be negative in many cases. It does not even capture \emph{down} in the second example.

Thirdly, we find that LSTM+ATT made a number of errors on sentences with negation words:

1. \emph{I have experienced no problems, [works]$_{pos}$ as anticipated.}

2. \emph{[Service]$_{neg}$ not the friendliest to our party!}

LSTMs typically only show their potential on large datasets. Without sufficient training examples, it may not be able to effectively capture various sequential patterns. Pretraining the network on larger document-level corpus addresses this problem.

Lastly, PRET+MULT makes fewer errors on recognizing neutral instances. This can also be observed from the macro-F1 scores in Table \ref{model comparison}. The lack of training examples makes the prediction of neutral instances very difficult for all previous methods. Knowledge from document-level examples with balanced labels compensates for this disadvantage.

\section{Conclusion}
The effectiveness of existing aspect-level neural models is limited due to the difficulties in obtaining training data in practice.
Our work is the first attempt to incorporate knowledge from document-level corpus for training aspect-level sentiment classifiers. We have demonstrated the effectiveness of our proposed approaches and analyzed the major benefits brought by the knowledge transfer. 
The proposed approaches can be potentially integrated with other aspect-level neural models to further boost their performance.

\bibliography{acl2018}
\bibliographystyle{acl_natbib}


\end{document}